\documentclass[conference]{IEEEtran}
\IEEEoverridecommandlockouts

\usepackage[T1]{fontenc}
\usepackage[english]{babel}
\usepackage{textcomp}
\usepackage[dvipsnames]{xcolor}
\usepackage{xspace}
\usepackage{graphicx}
\usepackage{amsfonts}
\usepackage{amsmath}
\usepackage{booktabs}
\usepackage[acronym]{glossaries}
\usepackage[hidelinks, bookmarksopen=true]{hyperref}
\usepackage[capitalize,noabbrev]{cleveref}
\usepackage{tikz}
\usepackage{multirow}
\usepackage{colortbl}
\usepackage{siunitx}
\usepackage{algpseudocode}
\usepackage{algorithm}
\usepackage{tabularx}
\usepackage{caption}
\usepackage[framemethod=default]{mdframed}
\usepackage{enumitem}
\usepackage{pifont}
\usepackage{fontawesome5}
\usepackage{tcolorbox}
\usepackage{makecell}

\glsdisablehyper
\AtBeginDocument{}
\tcbuselibrary{skins}

\def\ours{ChronosAD\xspace}

\makeatletter
\DeclareRobustCommand\onedot{\futurelet\@let@token\@onedot}
\def\@onedot{\ifx\@let@token.\else.\null\fi\xspace}

\makeatother

\definecolor{red}    {HTML}{b7211f}
\definecolor{orange} {HTML}{FFA500}
\definecolor{blue}   {HTML}{4169E3}
\definecolor{green}  {HTML}{147546}
\definecolor{purple} {HTML}{92268F}

\title{
ChronosAD: Leveraging Time Series Foundation Models for Accurate Anomaly Detection
\thanks{This work is supported by the project ``SUPREME: Sistemi Umano centrici per Prodotti e pRocEssi Manufatturieri Evoluti'' [PR Veneto FESR 2021-2027, Action 1.1.1, DGR 792, CUP D19J24000810007].}}

\author{
\IEEEauthorblockN{
Uzair Khan\IEEEauthorrefmark{1},
Luigi Capogrosso\IEEEauthorrefmark{2},
Francesco Biondani\IEEEauthorrefmark{1},
Michele Magno\IEEEauthorrefmark{3}\IEEEauthorrefmark{2},\\
Franco Fummi\IEEEauthorrefmark{1},
Francesco Setti\IEEEauthorrefmark{1},
Marco Cristani\IEEEauthorrefmark{1}}
\IEEEauthorblockA{
\IEEEauthorrefmark{1}University of Verona,
\IEEEauthorrefmark{2}Interdisciplinary Transformation University of Austria,
\IEEEauthorrefmark{3}ETH Zurich}
}

\begin{document}

\AddToHookNext{shipout/foreground}{%
\begin{tikzpicture}[overlay, remember picture]
    \node at ([yshift=-1cm]current page.north) {
        \normalsize\textcolor{gray}{This paper has been accepted for publication at the}
    };
    \node at ([yshift=-1.5cm]current page.north) {
        \normalsize\textcolor{gray}{24th IEEE International Conference on Industrial Informatics (INDIN), Melbourne, Australia, 2026}
    };
\end{tikzpicture}%
}

\maketitle

\bstctlcite{IEEEexample:BSTcontrol}

%%%%%%%%%% ABSTRACT.
\begin{abstract}
Time series anomaly detection is a crucial task in various domains, including finance, healthcare, and industry.
However, existing methods often struggle to generalize across different datasets, especially when anomalies are subtle or context-dependent.
To solve this issue, we introduce \ours{}, a novel architecture for anomaly detection that uses a time series foundation model as a feature extractor.
Specifically, it employs a two-stage pipeline: first, it uses the foundation model to extract embeddings for each time series in a zero-shot manner.
Then, a custom-developed Temporal Block, composed of Bidirectional Long Short-Term Memory (BiLSTM) and Multi-Head Attention, refines these embeddings to capture temporal dependencies and highlight salient patterns.
Unlike previous approaches, our model requires minimal task-specific tuning and demonstrates robust generalization across a wide range of domains, including industrial, medical, cyber-physical, and automotive systems.
Extensive experiments on 11 benchmarks show that \ours{} outperforms existing methods by 4.72\% in AUC and 6.60\% in AP on average.
The source code is available at \url{https://github.com/intelligolabs/ChronosAD}.
\end{abstract}

\glsresetall

%%%%%%%%%% BODY TEXT.
\section{Introduction} \label{cha:intro}

Anomaly detection is a critical process in various domains, including finance, cybersecurity, healthcare, and industry \cite{Chandola2009}.
It involves identifying data instances that deviate significantly from an expected pattern, often signaling fraudulent activities, medical conditions, or system malfunctions \cite{Khan2026}.

Recent advances in machine learning have led to the development of more sophisticated anomaly detection techniques, particularly in the domain of time series data \cite{Khan2026}.
However, these frameworks are typically tailored to specific domains \cite{Tang2023} and rely on handcrafted features or domain-specific heuristics, which limit their applicability and scalability.
This introduces significant challenges when deploying anomaly detection systems in diverse environments, as each domain may require a tailored model or considerable fine-tuning.

In recent years, computer vision-based foundation models have shown remarkable performance on downstream tasks like segmentation \cite{Li2025}, classification \cite{Simeoni2025}, and anomaly detection \cite{Damm2025} across various domains, such as healthcare \cite{Huang2024} and autonomous driving \cite{Audinys2025}.
Following this trend, several foundation models specifically tailored for time series have also recently emerged \cite{Garza2023,Ansari2024}.
These models, pre-trained on various multi-domain datasets, learn rich and generalizable representations that can be transferred to downstream tasks with minimal fine-tuning.
However, the adoption of foundation models for downstream time series tasks such as anomaly detection has not been fully explored.

Current time series foundation models possess two serious limitations.
First, they are predominantly univariate \cite{Ye2024}, which is insufficient for detecting anomalies in complex systems with multiple interdependent variables; in such settings, it is essential to model the cross-channel (or inter-variable) relationships alongside temporal dynamics.
Second, they are designed primarily for signal forecasting \cite{Ye2024}, but their forecasting performance remains limited in challenging settings \cite{Mulayim2024}.

In this paper, we propose \ours{}, a novel architecture that successfully uses a time series foundation model as a zero-shot feature extractor for anomaly detection across multiple domains.
It works end-to-end, adopting a two-stage pipeline.
In the first stage, it uses Chronos \cite{Ansari2024} to generate robust and task-agnostic embeddings for each univariate channel of a multivariate time series.
In the second stage, a custom Temporal Block, comprising Bidirectional Long Short-Term Memory (BiLSTM) networks and Multi-Head Attention, refines these embeddings to model temporal dependencies.

We conducted extensive experiments on 11 multi-domain datasets against 11 state-of-the-art methods, showing that \ours{} outperforms existing methods by 4.72\% in AUC and 6.60\% in AP on average.
We frame this work as a preliminary study on adapting time series foundation models for the downstream task of anomaly detection, adopting a supervised setting to facilitate the initial evaluation.
Given that this task is predominantly treated in an unsupervised manner, these results serve to validate our approach as a viable path for a more rigorous future investigation.
\section{Related Work} \label{cha:related}

%%%%%%%%%%
\emph{\textbf{Time Series Anomaly Detection.}} \label{ssec:ts_ad}
Recent research in time series anomaly detection has utilized a range of sophisticated deep learning models, including Multi-Layer Perceptrons (MLPs), Convolutional Neural Networks (ConvNets), LSTM networks, and Generative Adversarial Networks (GANs).

MLPs, valued for their straightforward design, were some of the initial models tested for these tasks.
These models can be effectively used to distinguish between normal and anomalous data points \cite{Mammadov2021}.
Nevertheless, they are unable to model sequential relationships, which frequently results in lower classification accuracy in complex real-world scenarios.

By applying one-dimensional convolutional filters, ConvNets have demonstrated remarkable success in identifying patterns and anomalies within time series data.
These filters integrate past observations to determine the current state of a signal \cite{Canizo2019,Kiranyaz2020}.
Although 1D ConvNets excel at identifying local temporal patterns, they are often inadequate for modeling relationships over extended periods, which is essential for anomalies characterized by long-term dependencies.

The combination of ConvNets with LSTM models to predict time series anomalies has been extensively investigated in \cite{Liu2020}.
Furthermore, standard LSTM-based anomaly detectors have been applied across various fields, including IoT networks \cite{Feng2021,Ullah2022,Shanmuganathan2023}, medicine \cite{Chauhan2015}, and industry \cite{Lee2018}.
Although Recurrent Neural Network (RNN) models are proficient in learning sequential patterns, they are susceptible to the vanishing gradient problem.
This issue complicates their ability to learn long-term dependencies, particularly in very long time series.

Generative Adversarial Networks (GANs) are also widely used for anomaly detection in time series.
In this framework, the generator learns to model the normal data distribution, and the discriminator identifies any deviations from this model \cite{Zenati2018,Schlegl2019,Zhou2019,Audibert2020}.
However, GANs often struggle to capture the complex temporal dynamics and long-range patterns inherent in time series data.
This limitation can lead to false positives or missed detections, particularly for rare or subtle anomalies that develop over time.
Furthermore, the training process for GANs can be unstable, which complicates the creation of a robust and reliable model, especially when dealing with highly dynamic or noisy data \cite{Zhang2022}.

%%%%%%%%%%
\emph{\textbf{Time Series Foundation Models.}} \label{ssec:ts_foundation_models}
Recent advances in foundation models have revolutionized time series analysis, substantially improving performance in a wide range of downstream tasks.
In \cite{Garza2023}, the authors introduced TimeGPT-1, the first foundation model specifically designed for time series analysis.
Using the GPT architecture \cite{Radford2018}, it adopts a generative approach to forecasting, capturing complex temporal dependencies to predict future data points with high accuracy.
Building on this momentum, \cite{Goswami2024} proposed the MOMENT family of open-source models, which incorporate large-scale pre-training and a dedicated benchmark for evaluation under limited supervision.
In another contribution, \cite{Rasul2024} developed a foundation model that integrates lagged variables as covariates and uses probabilistic forecasting techniques.
This enables the model to generate a distribution of plausible future outcomes, enhancing its capacity to capture uncertainty and variability in temporal data.
Similarly, \cite{Ansari2024} introduced Chronos, a cutting-edge foundation model for time series forecasting.
Chronos not only exceeds existing methods in performance but also exhibits strong zero-shot generalization across unseen datasets, underscoring its versatility and potential to streamline forecasting workflows.
Although these models perform well on both trained and unseen datasets, many of them are primarily designed for univariate settings.
This can limit their ability to explicitly capture interdependencies between variables, which are often relevant in multivariate anomaly detection scenarios.
To address this limitation, \ours{} adopts a multi-path architecture with channel fusion, enabling the model to aggregate information across multiple data streams and to model cross-channel interactions.
\section{Method} \label{cha:method}

\begin{figure}[!t]
    \centering
    \includegraphics[width=\linewidth]{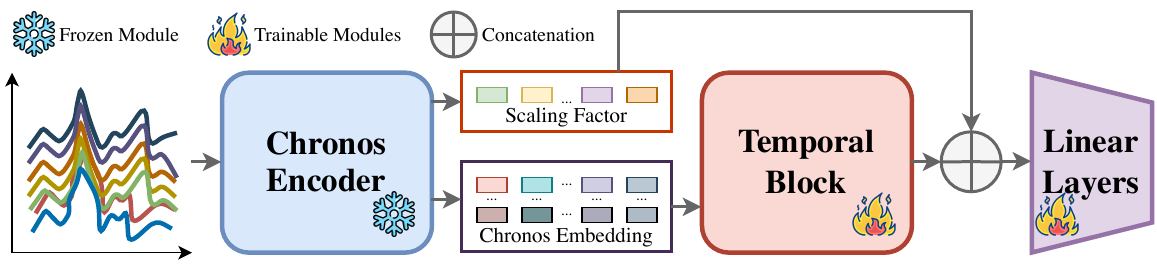}
    \caption{The \ours{} architecture.}
    \label{fig:fig_chronosad_architecture}
\end{figure}

In this section, we introduce \ours{}.
The overall architecture is illustrated in Figure \ref{fig:fig_chronosad_architecture}.

%%%%%%%%%%%%%%%%%%%%%%%%%%%%%%
\subsection{Problem Statement}
Let the features of the time series be denoted by $X\in\mathbb{R}^{C\times L_{total}}$, where $L_{total}$ represents the total length of the time series and $C$ is the total number of channels.
Throughout the rest of this article, the terms channels and variables will be used interchangeably.
The class of time series at a particular timestamp $t$ is represented by $y_t \in \{0, 1\}$, where 0 and 1 correspond to the normal and abnormal classes, respectively.

The time series is divided into non-overlapping segments, each of length at most $l$, referred to as the context length.
The context length determines the temporal window through which the model observes patterns to detect anomalies.
If any segment is shorter than the length $l$, it is padded with zeros at the end.
This segmentation results in $K$ segments, each containing $l$ consecutive timestamps after padding.
For the $n$-th segment $X_n \in \mathbb{R}^{C \times l}, \text{where }  n \in \{0, \dots, K-1\}$, let $y_n = \{y_i \mid i \in [t, t+l) \}$ represent the classes within that segment.
Following standard protocols \cite{Khan2026}, if at least one point is anomalous, the entire window is considered anomalous.

The model uses a threshold $\lambda$ to classify the anomaly score calculated from the segment's features.
As a result, the classification of anomalies can be expressed as a scoring function $f: \mathbb{R}^{C \times l} \to \mathbb{R}$ defined as follows:
\begin{equation}
    \hat{y}_n = \begin{cases}
    1, & \text{if } f(X_n) \geq \lambda \\
    0, & \text{otherwise}
    \end{cases}\;,
\end{equation}
where $\hat{y}_n \in \{0, 1\}$ represents the output class from \ours{}, with $\hat{y}=0$ indicating a normal segment and $\hat{y}=1$ indicating an abnormal segment.

Let $X \in \mathbb{R}^{N \times C \times l}$ represent the input of the time series, where $N$ is the batch size.
For each channel $c$ where $c=\{0,\dots,C-1\}$, the input is processed through the following pipeline.

%%%%%%%%%%%%%%%%%%%%%%%%%%%%%%
\subsection{ChronosAD Training Procedure} \label{sec:sec_training}

%%%%%%%%%%
\emph{\textbf{Chronos Encoder.}}
The time series of each channel $X_{c} \in \mathbb{R}^{N\times l}$ is an input to the same instance of Chronos \cite{Ansari2024}, which outputs a fixed-dimensional embedding vector:
\begin{equation}
    E_{c} = \text{Encoder}(X_{c})\;, \quad E_{c} \in \mathbb{R}^{N\times l \times d}\;,
\end{equation}
where $d$ is the embedding dimension.

Furthermore, the Chronos encoder provides a scaling factor $s$, where $s \in\mathbb{R}^{N}$.
For each instance in the batch, the scaling factor is calculated as the mean of the absolute values of the time series for a single channel $X_{c}$ in a context window of length $l$.
This is weighted by the attention mask $A$, which indicates valid time steps within the window to avoid contributions from padding timestamps.
Formally, the scaling factor is defined as:
\begin{equation}
    s = \frac{1}{\sum\limits_{j=0}^{l} A_j} \sum\limits_{j=0}^{l} \left(|X_{c,j}|A_j\right)\;, \quad \text{with } s > 0\;,
\end{equation}
where $X_{c,j}$ is the value of the time series for channel $c$ at index $j$, and $A_j$ is a binary mask that indicates whether the $j$-th time step is valid within the same context window.

The scaling factor plays a crucial role in describing the distribution of the time series within the specified context length.
This factor reflects the natural scale of the data, enabling the model to adapt its processing to varying magnitudes across different time series, thereby improving its ability to generalize across datasets with diverse characteristics.

Before feeding the embeddings into the Temporal Block, L2 normalization is applied to standardize the embedding magnitudes and mitigate the impact of extreme values.
This normalization step ensures numerical stability and promotes balanced input representations.

Next, the Scaled Exponential Linear Unit (SeLU) activation function is employed to non-linearly transform the embeddings, effectively clamping excessively negative values and shifting them toward a more neutral range.
This transformation helps to preserve gradient flow during backpropagation, thus mitigating the problem of vanishing gradients and enabling more efficient learning.
Specifically, let $\phi:R^{l\times d} \to R^{l\times d}$ be the SeLU activation function with $p = (0,\dots,l-1)$ and $v = (0,\dots,d-1)$, then:
\begin{equation}
    U_{c} = \phi\bigg(\frac{E_{c,p,v}}{
    \sqrt{\sum_{p=0}^{l-1}{E_{c,p,v}}}
    }\bigg)\;.
\end{equation}

%%%%%%%%%%
\emph{\textbf{Temporal Block.}}
\begin{figure}[t!]
    \centering
    \includegraphics[width=\linewidth]{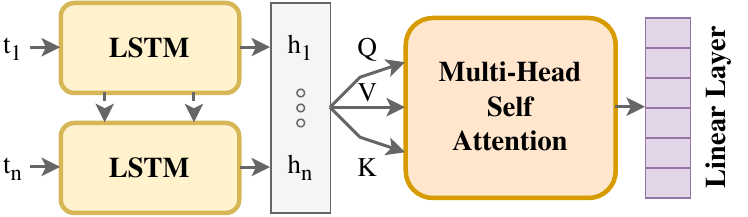}
    \caption{The custom Temporal Block used in \ours{} to capture temporal dependencies in sequence data.}
    \label{fig:fig_temporal_block_architecture}
\end{figure}
The embeddings $U_{c} \in \mathbb{R}^{N \times l \times d}$ are then passed into the Temporal Block (depicted in Figure \ref{fig:fig_temporal_block_architecture}), which processes the sequence data to capture temporal dependencies. 
The Temporal Block consists of two primary components: (1) a set of BiLSTM networks and (2) a Multi-Head Attention mechanism.
Each time step $t$ of the sequence is processed through a BiLSTM to encode forward and backward temporal features.
Formally, for a sequence:
\begin{equation}
    h_{c} = \text{BiLSTM}(U_{c})\;, \quad h_{c} \in \mathbb{R}^{N \times l \times 2r}\;,
\end{equation}
where $h_{c}$ represents the output of the LSTM for each channel $c$, and $r$ is the output dimension of the LSTM in each direction.
The resulting output sequence $h_{c}$ captures bidirectional temporal information.

These encoded temporal features are further refined using the Multi-Head Attention mechanism, which computes the attention-weighted representation of the sequence.
For each head, the queries ($Q$), keys ($K$), and values ($V$) are derived from $h_{c}$ as:
$Q = h_{c}W_Q$, $K = h_{c}W_K$, and $V = h_{c}W_V$, where $W_Q, W_K, W_V \in \mathbb{R}^{2r \times d_a}$ are learnable weight matrices, and $d_a$ is the dimensionality of each head of attention.

The attention weights are computed using the scaled dot-product mechanism:
\begin{equation}
   \alpha = \text{softmax}\left(\frac{QK^\top}{\sqrt{d_a}}\right)\;, \quad \alpha \in \mathbb{R} ^{N\times l \times l}\;,
\end{equation}
and the output of the attention head is given by: $z = \alpha V$, with $z \in \mathbb{R}^{N\times l \times d_a}$.

Each head is averaged, and a global sum operation, denoted by the function $\gamma: R^{N\times l \times d_a} \to R^{N \times l}$, is applied to reduce the dimensionality of $z$.
\begin{equation}
    \gamma(z)_{n,l} = \sum_{k = 1}^{d_a}{z_{n,l,k}}\;, \quad a_c = \gamma(z)\;,
\end{equation}
Then, a Linear Layer transforms $a_c$ into a fixed-dimensional representation $e_c = \text{Linear}(a_c)$, with $e_c \in \mathbb{R}^{N \times d_{e}}$, where $e_c$ is the output embedding that captures both temporal dependencies and attention-enhanced contextual information, and $d_e$ is the embedding dimension.
Finally, the scaling factor $s$ is concatenated at the end of the embeddings $e_c$: $\tilde{e}_c = \text{Concat}(e_c,s)$, with $\tilde{e}_c \in \mathbb{R}^{N \times (d_e + 1)}$.

After processing the embeddings through the Temporal Block, the output $\tilde{e}$ for each channel is concatenated along the feature dimension.
Specifically, for $C$ channels, the concatenated vector $E_{concat}$ is obtained as:
\begin{align}
    & E_{concat}=\text{Concat}(\tilde{e}_{1},\tilde{e}_{2},\dots,\tilde{e}_{C})\;, \\
    & E_{concat}\in\mathbb{R}^{N\times ( C\times (d_e + 1))}\;.
\end{align}

The concatenated embeddings $E_{concat}$ are then passed through a Fully Connected Network (FCN).
The FCN maps the concatenated features to a final prediction, which classifies the sequence as normal or abnormal.
Formally, the FCN output is given by $\hat{y}=\text{FCN}(E_{concat})$, with $R^{N}$, where $\hat{y}$ is the predicted class label, with $0$ indicating a normal sequence and $1$ indicating an abnormal sequence.

The output of the fully connected network, $\hat{y}$, is then used to compute the weighted Binary Cross-Entropy (WBCE) loss for training.
\section{Experiments} \label{cha:experiments}

%%%%%%%%%%
\emph{\textbf{Evaluation Metrics.}}
To evaluate \ours{}, we consider two key evaluation metrics: the Area Under the Receiver Operating Characteristic Curve (AUC) and the Average Precision (AP), following the evaluation protocol of \cite{Huang2023}.

%%%%%%%%%%
\emph{\textbf{Datasets.}}
For our experiments, we used a diverse collection of datasets, including seven time series datasets from the UCR Archive \cite{Dau2019}, a real-world bearing dataset (CWRU) \cite{Smith1997}, an ECG arrhythmia dataset (MIT-BIH) \cite{Moody2001}, a water treatment dataset (SWaT) \cite{Goh2017}, and a simulated dataset (Waveform) \cite{Zimek2013}.
These datasets were chosen to evaluate the performance of the proposed method across varying characteristics such as the number of samples, dimensions, classes, and sampling rates (Hz).

%%%%%%%%%%
\emph{\textbf{Competitors.}}
To assess the performance of \ours{}, we compared it with the following eleven methods: AnoGAN \cite{Schlegl2017}, ALAD \cite{Zenati2018}, Deep-SVDD \cite{Ruff2018}, BeatGAN \cite{Zhou2019}, GOAD \cite{Bergman2020}, USAD \cite{Audibert2020}, TLKF \cite{Shi2021}, GTA \cite{Chen2021}, NSIBF \cite{Feng2021}, ECOD \cite{Li2022}, and KalmanAE \cite{Huang2023}.
Since the current literature for this task is dominated by unsupervised approaches, we compared \ours{} with these baselines.
We frame this comparison to demonstrate that, in a scenario where labeled data can be acquired, our approach achieves high precision and robustness, overcoming the practical limitations and high false-positive rates typical of unsupervised methods.
While labeled anomalies are traditionally scarce, recent advances in digital twin technologies have made their acquisition increasingly feasible \cite{Biondani2025}.
Therefore, we position \ours{} as a complementary approach that can use such data when available, rather than a replacement for unsupervised methods.

%%%%%%%%%%
\emph{\textbf{Implementation Details.}}
The experiments were conducted using PyTorch 1.12 with CUDA version 11.2 on a system equipped with an RTX 4090 GPU.
For most datasets, the model was trained with a batch size of 32 to maintain consistency and computational efficiency.
However, for the MIT-BIH and SWaT datasets, a batch size of 128 was used to reduce variance during training, given their large sizes and the need for stable gradients.
Similarly, for all datasets, a learning rate of 0.001 was used, except for SWaT, where a lower learning rate of 0.0001 was used to ensure stable convergence due to the higher number of training epochs (1000) and the complex nature of the dataset.

%%%%%%%%%%%%%%%%%%%%%%%%%
\subsection{Quantitative Results}
\begin{table*}[t!]
    \centering
    \caption{Anomaly detection performance.
    All values are percentages (\%).
    In \textbf{bold}, the best results. 
    \underline{Underlined} the second best.}
    \begin{scriptsize}
        \resizebox{\textwidth}{!}{
            \begin{tabular}{l|cc|cc|cc|cc}
\multirow{2}{*}{Dataset} & \multicolumn{2}{c|}{\textbf{PPOC}~\cite{Dau2019}} & \multicolumn{2}{c|}{\textbf{TwoLeadECG}~\cite{Dau2019}} & \multicolumn{2}{c|}{\textbf{Strawberry}~\cite{Dau2019}} & \multicolumn{2}{c}{\textbf{TwoPatterns}~\cite{Dau2019}}\\
\cmidrule(lr){2-9}
& \textbf{AUC $\uparrow$} & \textbf{AP $\uparrow$} & \textbf{AUC $\uparrow$} & \textbf{AP $\uparrow$} & \textbf{AUC $\uparrow$} & \textbf{AP $\uparrow$} & \textbf{AUC $\uparrow$} & \textbf{AP $\uparrow$}\\
\toprule
AnoGAN~\cite{Schlegl2017}   & 52.30 & 71.88 & 52.04 & 73.76 & 61.97 & 74.36 & 51.46 & 88.89 \\
ALAD~\cite{Zenati2018}      & 50.39 & 67.07 & 51.64 & 72.80 & 52.93 & 69.69 & 50.50 & 88.23 \\
Deep-SVDD~\cite{Ruff2018}   & 65.73 & 82.08 & 85.56 & 89.82 & 72.80 & 73.74 & 83.61 & 97.03 \\
BeatGAN~\cite{Zhou2019}     & 86.02 & 74.40 & 78.42 & 74.62 & 91.38 & 89.66 & 75.85 & 88.60 \\
GOAD~\cite{Bergman2020}     & 66.82 & 71.59 & 52.97 & 54.90 & 55.69 & 61.32 & 74.07 & 77.39 \\
USAD~\cite{Audibert2020}    & 74.09 & 84.49 & 67.37 & 80.55 & 63.45 & 68.73 & 85.50 & 97.61 \\
TLKF~\cite{Shi2021}         & 75.05 & 84.61 & 71.49 & 82.55 & 72.88 & 79.84 & 75.19 & 85.22 \\
GTA~\cite{Chen2021}         & 80.31 & 84.30 & 72.60 & 80.05 & 81.00 & 85.87 & 84.91 & 96.48 \\
NSIBF~\cite{Feng2021}       & 75.47 & 82.23 & 88.09 & 90.78 & 76.44 & 72.24 & 88.71 & 92.74 \\
ECOD~\cite{Li2022}          & 81.86 & 92.58 & 82.68 & 88.19 & 82.59 & 88.78 & 90.91 & 95.85 \\
KalmanAE~\cite{Huang2023}   & \underline{94.53} & \underline{96.37} & \underline{91.14} & \underline{94.12} & \underline{93.92} & \underline{92.45} & \underline{94.28} & \underline{99.02} \\
\textbf{\ours{} (ours)}               & \textbf{96.05 \scriptsize{\color{green}{(+1.52)}}}
                                      & \textbf{98.05 \scriptsize{\color{green}{(+1.68)}}}
                                      & \textbf{100   \scriptsize{\color{green}{(+8.86)}}}
                                      & \textbf{100   \scriptsize{\color{green}{(+5.88)}}}
                                      & \textbf{98.18 \scriptsize{\color{green}{(+4.26)}}}
                                      & \textbf{99.05 \scriptsize{\color{green}{(+6.60)}}}
                                      & \textbf{100   \scriptsize{\color{green}{(+5.72)}}}
                                      & \textbf{100   \scriptsize{\color{green}{(+0.98)}}} \\
\midrule
\midrule
\multirow{2}{*}{Dataset} & \multicolumn{2}{c|}{\textbf{UWaveGestureLibraryY}~\cite{Dau2019}} & \multicolumn{2}{c|}{\textbf{FordA}~\cite{Dau2019}} & \multicolumn{2}{c|}{\textbf{SmallKitchenAppliances}~\cite{Dau2019}} & \multicolumn{2}{c}{\textbf{CWRU}~\cite{Smith1997}}\\
\cmidrule(lr){2-9}
& \textbf{AUC $\uparrow$} & \textbf{AP $\uparrow$} & \textbf{AUC $\uparrow$} & \textbf{AP $\uparrow$} & \textbf{AUC $\uparrow$} & \textbf{AP $\uparrow$} & \textbf{AUC $\uparrow$} & \textbf{AP $\uparrow$}\\
\toprule
AnoGAN~\cite{Schlegl2017}   & 82.74 & 98.52 & 50.27 & 71.02 & 48.76 & 83.48 & 72.47 & 87.13 \\
ALAD~\cite{Zenati2018}      & 52.64 & 94.55 & 50.22 & 71.68 & 51.94 & 84.40 & 47.94 & 91.84 \\
Deep-SVDD~\cite{Ruff2018}   & 65.68 & 96.38 & 64.54 & 80.48 & 74.17 & 91.68 & 92.09 & 99.02 \\
BeatGAN~\cite{Zhou2019}     & 91.46 & 97.73 & 73.35 & 84.00 & 83.40 & 82.11 & 97.91 & 95.84 \\
GOAD~\cite{Bergman2020}     & 87.28 & 96.87 & 54.65 & 73.66 & 48.89 & 52.39 & 53.20 & 58.17 \\
USAD~\cite{Audibert2020}    & 90.06 & 99.30 & 63.81 & 79.83 & 62.31 & 85.96 & 85.86 & 97.88 \\
TLKF~\cite{Shi2021}         & 79.09 & 88.32 & 62.54 & 71.64 & 67.91 & 76.73 & 86.43 & 95.61 \\
GTA~\cite{Chen2021}         & 89.43 & 92.67 & 74.28 & 84.90 & 76.05 & 87.96 & 95.57 & 96.78 \\
NSIBF~\cite{Feng2021}       & 92.63 & 97.58 & 73.29 & 82.72 & 84.66 & 91.95 & 97.18 & 98.08 \\
ECOD~\cite{Li2022}          & 93.50 & 97.00 & 75.28 & 84.80 & 83.78 & 91.18 & 97.50 & 98.57 \\
KalmanAE~\cite{Huang2023}   & \underline{95.67} & \underline{99.59} & \underline{79.30} & \underline{87.64} & \underline{89.48} & \underline{93.32} & \underline{99.02} & \underline{99.68} \\
\textbf{\ours{} (ours)}               & \textbf{98.38 \scriptsize{\color{green}{(+2.71)}}}
                                      & \textbf{99.75 \scriptsize{\color{green}{(+0.16)}}}
                                      & \textbf{97.43 \scriptsize{\color{green}{(+18.13)}}}
                                      & \textbf{97.29 \scriptsize{\color{green}{(+9.65)}}}
                                      & \textbf{89.79 \scriptsize{\color{green}{(+0.31)}}}
                                      & \textbf{94.08 \scriptsize{\color{green}{(+0.76)}}}
                                      & \textbf{100   \scriptsize{\color{green}{(+0.98)}}}
                                      & \textbf{100   \scriptsize{\color{green}{(+0.32)}}}\\
\midrule
\midrule
\multirow{2}{*}{Dataset} & \multicolumn{2}{c|}{\textbf{MIT-BIH}~\cite{Moody2001}} & \multicolumn{2}{c|}{\textbf{SWaT}~\cite{Goh2017}} & \multicolumn{2}{c|}{\textbf{Waveform}~\cite{Zimek2013}} & \multicolumn{2}{c}{\textbf{Average}}\\
\cmidrule(lr){2-9}
& \textbf{AUC $\uparrow$} & \textbf{AP $\uparrow$} & \textbf{AUC $\uparrow$} & \textbf{AP $\uparrow$} & \textbf{AUC $\uparrow$} & \textbf{AP $\uparrow$} & \textbf{AUC $\uparrow$} & \textbf{AP $\uparrow$}\\
\toprule
AnoGAN~\cite{Schlegl2017}   & 75.33 & 58.75 & 61.78 & 77.43 & 52.91 & 34.89 & 60.18 & 74.55 \\
ALAD~\cite{Zenati2018}      & 71.66 & 62.07 & 77.52 & 87.02 & 68.35 & 35.50 & 56.88 & 74.98 \\
Deep-SVDD~\cite{Ruff2018}   & 85.76 & 66.64 & 87.04 & 88.65 & 84.68 & 45.12 & 85.33 & 82.78 \\
BeatGAN~\cite{Zhou2019}     & 92.99 & 89.87 & 89.98 & 89.99 & 83.19 & 44.27 & 88.36 & 82.62 \\
GOAD~\cite{Bergman2020}     & 72.70 & 87.61 & 63.16 & 70.49 & 77.08 & 37.49 & 64.21 & 67.44 \\
USAD~\cite{Audibert2020}    & 88.91 & 76.00 & 87.43 & 92.62 & 85.56 & 42.41 & 87.32 & 82.30 \\
TLKF~\cite{Shi2021}         & 80.34 & 72.85 & 72.49 & 79.38 & 64.13 & 43.11 & 73.41 & 78.16 \\
GTA~\cite{Chen2021}         & 72.96 & 62.78 & 91.32 & 94.12 & 63.85 & 39.12 & 81.11 & 82.27 \\
NSIBF~\cite{Feng2021}       & 84.53 & 72.88 & 95.76 & 96.47 & 62.25 & 45.20 & 83.54 & 83.89 \\
ECOD~\cite{Li2022}          & 80.44 & 70.39 & 82.73 & 79.43 & 74.88 & 40.09 & 84.61 & 82.46 \\
KalmanAE~\cite{Huang2023}   & \underline{95.06} & \underline{91.48} & \underline{97.32} & \underline{98.63} & \underline{87.04} & \underline{46.15} & \underline{92.43} & \underline{90.76} \\
\textbf{\ours{} (ours)}               & \textbf{97.59 \scriptsize{\color{green}{(+2.53)}}}
                                      & \textbf{96.95 \scriptsize{\color{green}{(+5.47)}}}
                                      & \textbf{98.84 \scriptsize{\color{green}{(+1.52)}}}
                                      & \textbf{98.81 \scriptsize{\color{green}{(+0.16)}}}
                                      & \textbf{92.41 \scriptsize{\color{green}{(+5.37)}}}
                                      & \textbf{86.93 \scriptsize{\color{green}{(+40.78)}}}
                                      & \textbf{97.15 \scriptsize{\color{green}{(+4.72)}}}
                                      & \textbf{97.36 \scriptsize{\color{green}{(+6.60)}}}\\
\bottomrule
\end{tabular}
        }
    \end{scriptsize}
    \label{tab:chronosad_results}
\end{table*}

The results are presented in Table \ref{tab:chronosad_results}.
On UCR datasets such as TwoPatterns, UWaveGestureLibraryY, and FordA, we achieve AUC values close to or at 100\%.
For example, our model's flawless performance on TwoPatterns and near-optimal results on UWaveGestureLibraryY underscore its precision in detecting nuanced deviations across synthetic and gesture-based time series.
Similarly, its dominance in FordA, a dataset oriented toward automotive fault detection, further underscores its relevance for applied diagnostic tasks.

Real-world datasets such as CWRU and SWaT further validate the robustness of our model.
Specifically, on CWRU, the model achieves perfect scores, indicating its ability to precisely identify faults in industrial bearing systems.
For SWaT, a challenging multivariate dataset representing cyber-physical systems, \ours{} achieves the highest score among the compared methods, indicating that the model effectively captures inter-channel dependencies.
Similarly, \ours{} excels in detecting anomalies in MIT-BIH ECG recordings, a critical requirement in medical diagnostics.

In summary, \ours{} achieves consistent performance across benchmarks, with an average AUC of 97.15\% and an AP of 96.36\%.
These results highlight the potential of time series foundation models for anomaly detection.
Future work will focus on extending the approach to unsupervised settings to enable more direct comparisons with existing methods.

%%%%%%%%%%%%%%%%%%%%%%%%%
\subsection{Ablation Study}
\begin{table*}[t!]
    \centering
    \caption{Ablation study across MIT-BIH, UWaveGestureLibraryY, and TwoLeadECG, as well as the overall average performance across all datasets.
    The performance drops compared to the baseline are indicated in \textcolor{red}{red}.}
    \begin{scriptsize}
    \resizebox{\textwidth}{!}{
        \begin{tabular}{l|cc|w{c}{2.0cm} w{c}{2.0cm}|cc|w{c}{2.2cm} w{c}{2.2cm}}
\toprule
\textbf{Ablation} & \multicolumn{2}{c|}{\textbf{MIT-BIH}~\cite{Moody2001}} & \multicolumn{2}{c|}{\textbf{UWaveGestureLibraryY}~\cite{Dau2019}} & \multicolumn{2}{c|}{\textbf{TwoLeadECG}~\cite{Dau2019}} & \multicolumn{2}{c}{\textbf{Average}} \\
\cmidrule(lr){2-9}
 & \textbf{AUC $\uparrow$} & \textbf{AP $\uparrow$} & \textbf{AUC $\uparrow$} & \textbf{AP $\uparrow$} & \textbf{AUC $\uparrow$} & \textbf{AP $\uparrow$} & \textbf{AUC $\uparrow$} & \textbf{AP $\uparrow$} \\
\toprule
Removing Chronos        & 97.17 & 96.19 & 8.29 & 73.44 & 99.51 & 99.49 & 78.83 \textcolor{red}{(-18.34)} & 87.60 \textcolor{red}{(-9.76)} \\
Removing L2             & 96.34 & 97.16 & 67.21 & 90.61 & 99.98 & 99.98 & 93.23 \textcolor{red}{(-3.94)} & 95.72 \textcolor{red}{(-1.64)} \\
Removing SeLU           & 97.35 & 96.69 & 98.52 & 99.77 & 100.00 & 100.00 & 96.82 \textcolor{red}{(-0.35)} & 97.04 \textcolor{red}{(-0.32)} \\
Removing Attention      & 96.65 & 95.67 & 95.62 & 99.16 & 99.98 & 99.98 & 95.77 \textcolor{red}{(-1.40)} & 96.46 \textcolor{red}{(-0.90)} \\
Removing BiLSTM         & 97.11 & 96.50 & 97.93 & 99.57 & 99.99 & 99.99 & 96.97 \textcolor{red}{(-0.20)} & 97.13 \textcolor{red}{(-0.23)} \\
Removing Scaling Factor & 97.58 & 96.91 & 98.29 & 99.70 & 100.00 & 100.00 & 97.12 \textcolor{red}{(-0.05)} & 97.14 \textcolor{red}{(-0.22)} \\
\midrule
Baseline & 97.59 & 96.95 & 98.59 & 99.83 & 100.00 & 100.00 & 97.15 & 97.36 \\
\bottomrule
\end{tabular}}
    \end{scriptsize}
    \label{tab:ablation_study}
\end{table*}

The ablation study (Table \ref{tab:ablation_study}) highlights the role of each \ours{} component.
The Chronos encoder is the most critical component: removing it drops AUROC to 78.83\% and AP to 87.60\% ($\downarrow$18.34\%, $\downarrow$9.76\%).

Chronos embeddings vary in magnitude, which can lead to vanishing/exploding gradients in the LSTM.
Removing L2 normalization confirms this: AUROC and AP decrease by 3.94\% and 1.64\%, showing its role in stabilizing training.

Removing attention reduces AUROC to 95.77\% ($\downarrow$1.40\%) and AP to 96.46\% ($\downarrow$0.90\%).
Its benefit depends on the dataset's complexity (\emph{e.g.}, MIT-BIH and UWaveGestureLibraryY rely heavily on attention, while TwoLeadECG relies on it to a lesser extent).

Other components have smaller effects:
\begin{itemize}
    \item SeLU activation: AUROC $\downarrow$0.35\%, AP $\downarrow$0.32\% (helps gradient flow).
    \item BiLSTM: AUROC $\downarrow$0.20\%, AP $\downarrow$0.23\% (captures bidirectional dependencies).
    \item Scaling factor: AUROC $\downarrow$0.05\%, AP $\downarrow$0.22\%.
\end{itemize}

In general, the Chronos encoder, L2 normalization, and attention are the most influential components, while other modules refine robustness and performance.

%%%%%%%%%%%%%%%%%%%%%%%%%
\subsection{Latent Space Analysis}
\begin{figure*}[t!]
    \centering
    \includegraphics[width=.9\linewidth]{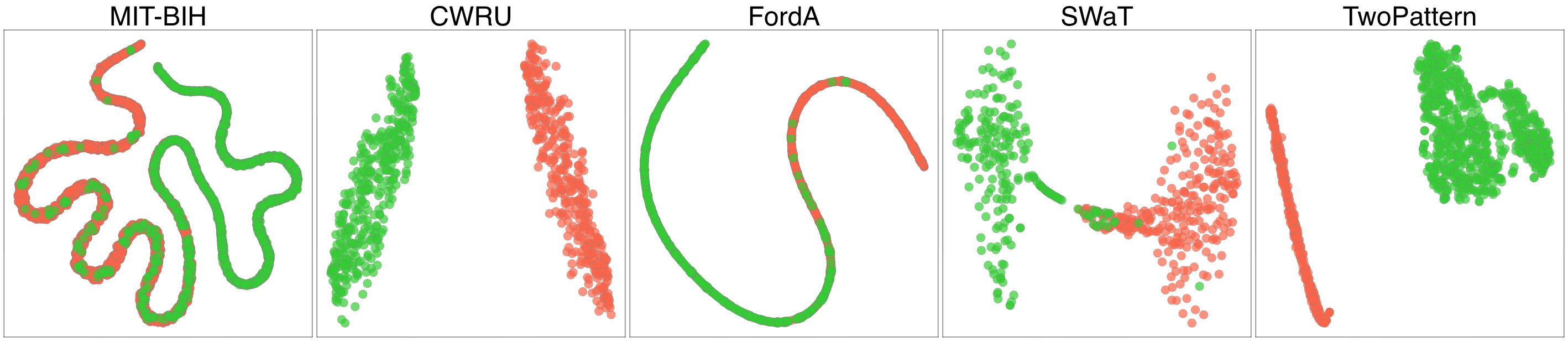}
    \caption{The visualization shows the latent space of the model using t-SNE for diverse datasets.
    The \textcolor{red}{red} shows the anomalies while the \textcolor{green}{green} shows the normal time series.}
    \label{fig:qualitative_tsne}
\end{figure*}

Figure \ref{fig:qualitative_tsne} shows the t-SNE projection of the latent space from the second-last layer of \ours{} using the test set.

In datasets such as CWRU, TwoPatterns, and SWaT, normal and anomalous points form distinct clusters with minimal overlap, indicating that the model captures discriminative features for anomaly detection.
The structured distribution suggests that the model organizes similar points along smooth manifolds, learning meaningful patterns.

In contrast, datasets such as MIT-BIH and FordA exhibit partial overlap between normal and anomalous points, creating transition regions where boundaries are less distinct.
This reflects the complexity of these datasets.
Despite this overlap, the latent space still shows a clear tendency to separate normal and anomalous series, demonstrating that the model learns representations that capture subtle differences while maintaining separability.
\section{Conclusions} \label{sec:conclusions}

In this paper, we introduce \ours{}, a novel and powerful two-stage deep learning architecture for time series anomaly detection.
Our approach leverages the Chronos time series foundation model as a zero-shot feature extractor and a custom-tailored Temporal Block to capture temporal dependencies and highlight salient patterns.
With consistent performance across diverse domains, from healthcare to industrial sensor data, \ours{} validates time series foundation models as a viable path for anomaly detection, paving the way for future unsupervised research.

%%%%%%%%%% BIBLIOGRAPHY.
\bibliographystyle{IEEEtran}
\bibliography{bibliography}

\end{document}